\documentclass[10pt,journal,compsoc]{IEEEtran}
\ifCLASSOPTIONcompsoc
  \usepackage[nocompress]{cite}
\else
  \usepackage{cite}
\fi
\ifCLASSINFOpdf
\else
\fi

\usepackage{graphicx}
\usepackage{algorithm}
\usepackage{algorithmic}

\usepackage{tikz}
\usepackage{comment}
\usepackage{amsmath} 
\usepackage{amsthm}
\usepackage{amssymb}
\usepackage{color}
\usepackage[accsupp]{axessibility}  
\usepackage{multirow}
\usepackage{subfig}
\usepackage{makecell}
\usepackage{pgfplots}
\usepackage{xspace}
\usepackage{booktabs}
\usepackage{tabularray}
\usepackage{svg}

\newcommand{\name}[0]{TRG\xspace}
\newcommand{\ie}{\textit{i}.\textit{e}.}

\newcommand{\etal}{\textit{et al. }}
\newcommand{\etc}{\textit{etc.}}

\usepackage{scalerel}
\usepackage{tikz}
\usetikzlibrary{svg.path}
\definecolor{orcidlogocol}{HTML}{A6CE39}
\tikzset{
    orcidlogo/.pic={
        \fill[orcidlogocol] svg{M256,128c0,70.7-57.3,128-128,128C57.3,256,0,198.7,0,128C0,57.3,57.3,0,128,0C198.7,0,256,57.3,256,128z};
        \fill[white] svg{M86.3,186.2H70.9V79.1h15.4v48.4V186.2z}
        svg{M108.9,79.1h41.6c39.6,0,57,28.3,57,53.6c0,27.5-21.5,53.6-56.8,53.6h-41.8V79.1z M124.3,172.4h24.5c34.9,0,42.9-26.5,42.9-39.7c0-21.5-13.7-39.7-43.7-39.7h-23.7V172.4z}
        svg{M88.7,56.8c0,5.5-4.5,10.1-10.1,10.1c-5.6,0-10.1-4.6-10.1-10.1c0-5.6,4.5-10.1,10.1-10.1C84.2,46.7,88.7,51.3,88.7,56.8z};
    }
}
\newcommand\orcidicon[1]{\href{https://orcid.org/#1}{\mbox{\scalerel*{
                \begin{tikzpicture}[yscale=-1,transform shape]
                \pic{orcidlogo};
                \end{tikzpicture}
            }{|}}}}
\usepackage{hyperref}
\hypersetup{hidelinks,colorlinks=true,allcolors=black,pdfstartview=Fit,breaklinks=true}
\begin{document}
%
\title{Knowledge Distillation via Token-level Relationship Graph}
%
%
%
%

\author{
        Shuoxi~Zhang,
        Hanpeng~Liu, 
        Kun He\IEEEauthorrefmark{1}${\textsuperscript{\orcidicon{0000-0001-7627-4604}}}$
        , \IEEEmembership{Senior~Member,~IEEE}
\IEEEcompsocitemizethanks{\IEEEcompsocthanksitem The authors are with the School
of Computer Science and Technology, Huazhong University of Computer Science and Technology, Wuhan,
Chine, 430074.\protect\\
\IEEEcompsocthanksitem 
The first two authors contributed equally.
Corresponding author: Kun He (E-mail: brooklet60@hust.edu.cn).
}
\thanks{Manuscript received June 15, 2023; revised ??, ????.}}

%
%

\markboth{
IEEE Transactions on Big Data,~Vol.~??, No.~?, August~20??}
{Zhang \MakeLowercase{\textit{et al.}}: Knowledge Distillation via Token-wise Graph}
%



\IEEEtitleabstractindextext{%
\begin{abstract}
Knowledge distillation is a 
powerful technique for transferring knowledge from a pre-trained teacher model to a student model. However, the true potential of knowledge transfer has not been fully explored. Existing approaches primarily focus on distilling individual information or instance-level relationships, overlooking the valuable information embedded in token-level relationships, which may be particularly affected by the long-tail effects. 
To address the above limitations, we propose a novel method called Knowledge Distillation with Token-level Relationship Graph (\name) that leverages the token-wise relational knowledge to enhance the performance of knowledge distillation. By employing \name, the student model can effectively emulate higher-level semantic information from the teacher model, resulting in improved distillation results. To further enhance the learning process, we introduce a token-wise contextual loss called contextual loss, which encourages the student model to capture the inner-instance semantic contextual of the teacher model. We conduct experiments to evaluate the effectiveness of the proposed method against several state-of-the-art approaches. Empirical results demonstrate the superiority of \name across various visual classification tasks, including those involving imbalanced data. Our method consistently outperforms the existing baselines, establishing a new state-of-the-art performance in the field of knowledge distillation. 
\end{abstract}

}

\maketitle

\IEEEdisplaynontitleabstractindextext

%
\IEEEpeerreviewmaketitle

\IEEEraisesectionheading{\section{Introduction}\label{sec:introduction}}


 

\IEEEPARstart{D}{eep} learning models have 
gained remarkable achievements across a wide range of 
applications in various fields, including computer vision~\cite{voulodimos2018deep}, natural language processing ~\cite{deng2018deep}, speech recognition~\cite{deng2014ensemble}, game playing~\cite{AlphaGo}, \etc\, However, the high computation and memory requirements of deep neural networks (DNNs) limit their practical deployment in resource-constrained environments such as mobile devices and embedded systems. 

To address this challenge, 
Knowledge Distillation (KD)~\cite{hinton2015distilling} has emerged as a promising research field. 
Drawing inspiration from the concept of transfer learning~\cite{pan2009survey},  KD 
utilizes a high-capacity and complex teacher network to guide the training process of a 
lightweight student network 
so as to transfer the teacher's knowledge. The knowledge can be transferred through two primary components: prediction logits~\cite{hinton2015distilling} and intermediate features~\cite{chen2021cross, zagoruyko2016paying, kim2018paraphrasing}. Logit-based distillation is based on the idea that the soft prediction probabilities contain more information than hard labels. By using these probabilities, the performance of the student model can be maintained effectively after the distillation process. On the other hand, feature-based distillation methods directly align the feature activations of the teacher network and the student network. This approach achieves the transfer of knowledge at the feature level, facilitating effective learning on the student model.


Recently, feature distillation methods have gained significant attention due to their ability to provide superior performance on various tasks compared to logit-based methods. The preference arises from the flexibility offered by feature representation in capturing rich information. Despite the outstanding achievements, both logit and feature distillation approaches face a notable challenge known as the capacity gap. This challenge stems from the disparity in 
scale between teacher and student models, resulting in a noticeable gap in their capacity even after the distillation process. Such capacity gap cannot be 
ignored, 
as it leads to 
the heavily rely of student model 
on the teacher model for effective knowledge transfer.

To tackle the issue of feature mismatch, Park \etal propose a novel approach to transfer knowledge utilizing the correlation information among instances~\cite{park2019relational}. The authors argue that instead of relying solely on individual knowledge distillation, the mutual relationship between instances can offer additional structured information to alleviate the capacity gap encountered in previous distillations. This groundbreaking contribution lays  the foundation for a new subfield within the KD paradigm, commonly referred to as relation-based distillation. 
The underlying concept 
is that the information extracted from pairs of instances should remain invariant, regardless of the architecture differences between networks of teacher and student.

\begin{figure*}

\centering 
\includegraphics[width=0.95\textwidth]{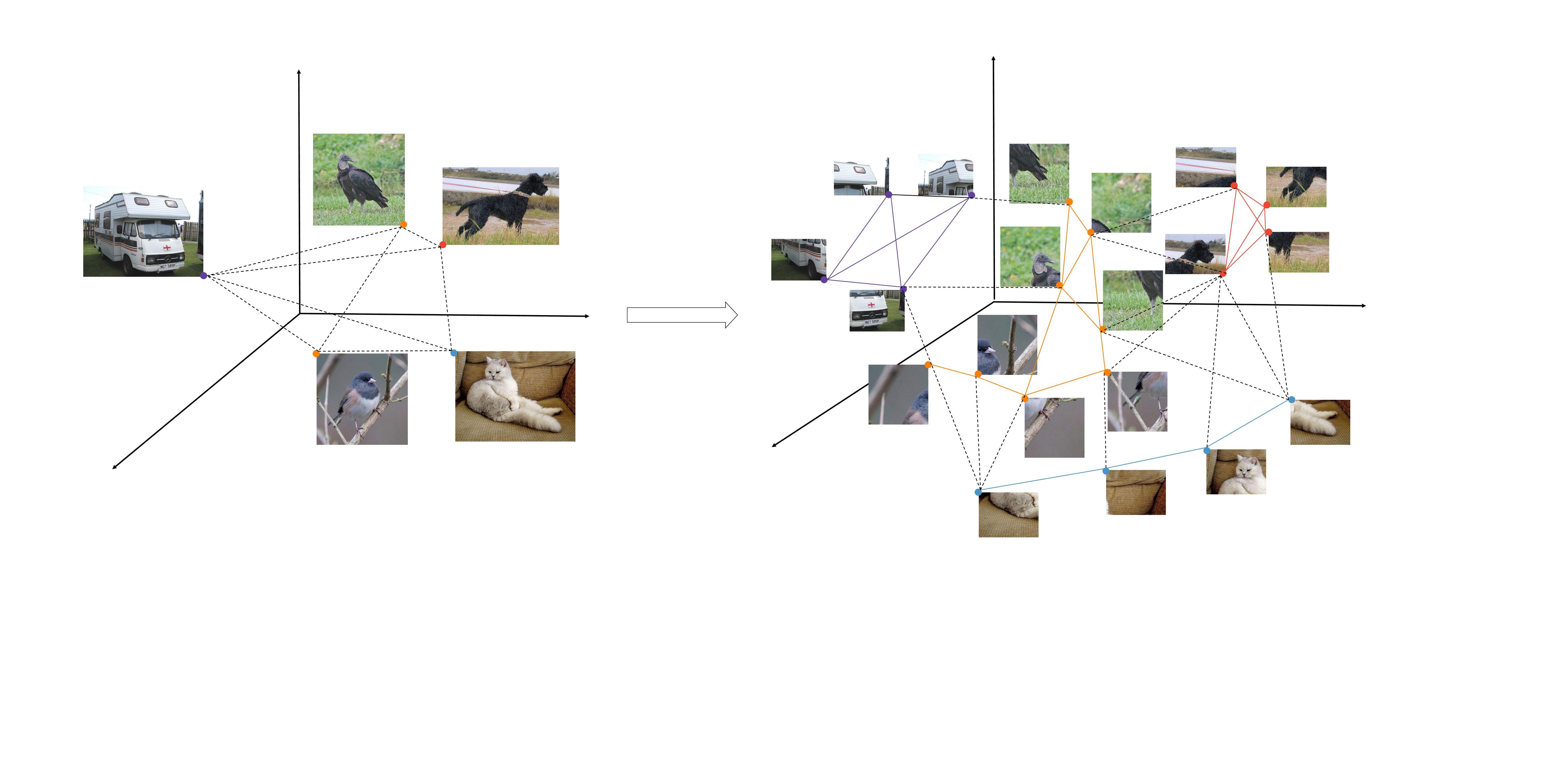}
\caption{
Illustration of the instance-level graph (left) and token-level graph (right).
The token-level graph  containing fine-grained information facilitates the knowledge distillation. For instance, the fur tokens for images with labels of cat or dog should share similar patterns.}
\label{fig:token-instance} 
\end{figure*}

Among various methods available for representing the structural information, graphs emerge as a prominent approach for capturing features and hierarchical relational patterns in a local connectivity manner. Consequently, graphs serve as suitable carriers for knowledge distillation. Graph-based distillation methods leverage the graph embedding~\cite{cai2018comprehensive, hamilton2017representation} or incorporate additional graph neural networks (GNNs)~\cite{scarselli2008graph} to effectively represent and transfer the relational knowledge during the KD process. Liu \etal~\cite{liu2019IRG} first propose to distill the instance-level relational information by constructing an instance relationship graph (IRG). Their approach distills graph-formed knowledge from the teacher model to the student model. Additionally,   HKD~\cite{zhou2021distilling} employs GNNs for the purpose of knowledge distillation.

However, we observe that 
existing graph-based distillation methods primarily focus on capturing relational knowledge at the \textit{instance level}, which may lose fine-grained information in preserving the semantics inherent to individual images. Therefore, relying solely on distillation at the instance level may lead to potential information loss during the KD process. Moreover, in scenarios with imbalanced data, the effectiveness of the instance-level relation-based KD paradigm may be compromised due to the \textit{long-tail effect} caused by imbalanced classes, introducing additional challenges. Consequently, 
there is an urgent need to further explore existing graph-based KD methods to overcome these limitations.

In Fig.~\ref{fig:token-instance}, we 
illustrate a comparison of the information captured by \textit{instance-level} graph and \textit{token-level} graph. The token-level graph may offer a more detailed analysis by considering the intricate patterns and characteristics exhibited by various patch tokens both within individual images and across different images. For instance, 
when recognizing images of \textit{cat} and \textit{dog},
it is expected that the fur patches of both animals should share some similar pattern. By incorporating these fine-grained information into the relationship graph, we can effectively transfer knowledge that complements the instance-level knowledge. 
Thus, utilizing a token-level graph to transfer knowledge may enhance the overall representation of relationships and similarities across various elements within and across images.

In this work, we propose a novel knowledge distillation method called \textbf{T}oken \textbf{R}elationship \textbf{G}raph (\name) for KD. Our approach is motivated by the notion that the relationships between tokens can capture the semantics within individual images as well as the semantic congruity across multiple images. 
Specifically, we partition the feature maps obtained from both teacher and student networks into fixed-sized patch tokens. These tokens then serve as the basis for constructing attributed graphs for each network, where each node represents a token and the node attributes correspond to the token embeddings. The graph edges are generated using the K-nearest neighbors (\textit{k}-NN) of each node, establishing a relationship graph within each mini-batch of training data samples.

Imbalanced data scenarios are often difficult to handle with individual KD methods owing to the long-tail effect. By focusing on the token-level relationships, our approach enables a more comprehensive understanding of the semantic content across images, even in the presence of imbalanced data. Additionally, to capture the contextual semantics within the instances, we also employ a contextual loss, aiming to preserve the semantics of individual instances.
 
The flexibility of our KD method is demonstrated by its compatibility with novel network architectures, particularly the Vision Transformers~\cite{dosovitskiy2020image} (ViTs), which are primarily based on patch tokenization. By incorporating the image-level token patching, we extend the applicability of our KD method to ViTs and other network architectures. To evaluate the effectiveness of \name, we conduct extensive experiments and do comparison against state-of-the-art baselines. Empirical results demonstrate that our method achieves state-of-the-art performance on several popular visual datasets, using various network architectures of both convolutional neural networks (CNNs) and ViTs. These findings validate the superiority of \name in distilling semantic knowledge and highlight its potential for improving performance in visual recognition tasks.

Our main contributions can be summarized as follows:
\begin{itemize}

    \item We propose a new method called Token Relationship Graph (\name) to efficiently distill the token-level relational knowledge across instances. To the best of our knowledge, this is the first work that proposes capturing the relational information among tokens rather than instances in the context of knowledge distillation.
    
    \item  We incorporate the contextual similarity loss between tokens within the individual instances. By leveraging the inner-instance contextual loss and intra-instance token-level graph, our method captures the semantics of individual instances and the relationships between images, enabling the effective conveyance of inherent relational knowledge. 
    
    \item 
    Extensive experiments, utilizing various model architectures and popular visual datasets, are conducted to  evaluate the superior performance of the proposed approach 
    over state-of-the-art baselines. 
\end{itemize}

\label{01intro}

\section{Related Work}
This section provides a brief review of related work on knowledge distillation, graph representation, and their combinations, meanwhile we also emphasize the distinctive aspects of our work in comparison to existing studies.

\subsection{Knowledge Distillation}
The knowledge distillation (KD) originally introduced by Hinton \etal~\cite{hinton2015distilling} aims to transfer knowledge from a teacher network to a student network by utilizing soft labels. While KD has contributed to network compression by using logit probability information, researchers argue that the rich information may be embedded in intermediate features. Romero \etal \cite{romero2014fitnets} first propose the feature-based distillation, which aims to transfer knowledge by aligning the intermediate features of teacher and student models. Such alignment allows the student to learn from more detailed information encoded by the teacher, resulting in improved performance. Additionally, Zagoruyko \etal \cite{zagoruyko2016paying} introduce attention transfer (AT), which enforces the student network to match the attention maps of a more powerful teacher network.

However, feature-based distillation methods often encounter challenges due to feature misalignment caused by the size difference between the teacher and student models. Researchers argue that knowledge distillation should transfer the structural knowledge learned by the teacher. In this context, Park \etal \cite{park2019relational} propose a pioneer approach of relation KD, where they consider the mutual relations of instances, providing structured information to address the feature misalignment problem. Their work establishes a new branch 
known as relation-based distillation.
Subsequently, Tung \etal \cite{tung2019similarity} introduce a knowledge distillation method focused on preserving similarity. Their approach involves transferring pairwise similarity knowledge to the student model. Similarly, Peng \etal \cite{peng2019correlation} propose a knowledge distillation method based on correlation congruence, which incorporates both instance-level information and correlations between instances in the distilled knowledge. By leveraging correlation congruence, the student network can effectively learn and capture the correlations between instances. These methods expand the scope of knowledge distillation by considering different aspects of structural and relational knowledge transfer.

\subsection{Graph-based Distillation}
Graphs, with their ability to capture and represent complex relationship information, are applied successfully in various domains such as social network analysis, recommendation systems, bioinformatics, and computer vision. The work of Sperduti \etal \cite{sperduti1997supervised} pioneer the application of neural network architectures to graphs, which subsequently motivates the exploration and development of Graph Neural Networks (GNNs). Building upon this foundation, several studies of Xu \etal \cite{xu2018powerful}, Hamilton \etal \cite{hamilton2017inductive}, and Velickovic \etal \cite{velivckovic2017graph} establish the paradigm of embedding graphs within neural network architectures.

Several works propose utilizing graphs to represent and convey the relational knowledge for knowledge distillation. The IRG method \cite{liu2019IRG} is the first to employ instance-wise graph structures to capture relational knowledge learned from the teacher model. Chen \etal \cite{chen2020learning} introduce a locality-preserving loss into the KD paradigm to preserve the relationships between samples in the teacher-student framework. Zhou \etal \cite{zhou2021distilling} utilize graph neural networks to integrate both individual and relational knowledge, enabling the preservation of valuable relationship information.

Existing graph-based distillation methods primarily emphasize the relational information among instances, potentially overlooking important image semantics. In this work, we propose a new approach that incorporates a token relationship graph within a batch-size data pipeline. Importantly, our method captures both inner-image semantics and intra-image relationships without incurring excessive computational cost. Also, our approach can be easily integrated into various popular deep neural network architectures through the addition of simple graph branches and token patching. 

\label{02RW}

\section{Preliminary}
In this section, we provide an overview of the notations 
relevant to our study, and the vanilla KD method originally proposed by Hinton \etal~\cite{hinton2015distilling}.

\subsection{Notations}
For $(\boldsymbol{x, y}) = \{(\boldsymbol{x}_i, y_i),  i=1,2, \cdots, N\}$, $\boldsymbol{x}_i$ is the i-th image and $\boldsymbol{y}_i$ is the corresponding label.
The $i$-th image 
$\boldsymbol{x}_i \in \mathbb{R}^{C \times H \times W}$, where $C$ is the number of channels, and $H$ and $W$ denote the height and width of the image, respectively. The features and outputs extracted from the model are denoted as $\mathbf{F}$ and $\boldsymbol{z}$, respectively. The logits predicted by the teacher and student networks are represented as $\boldsymbol{z}^{\mathcal{T}}$ and $\boldsymbol{z}^{\mathcal{S}}$, which are primarily used in individual knowledge distillation. The feature representations learned by the teacher and student networks are denoted as $\mathbf{F}^{\mathcal{T}}$ and $\mathbf{F}^{\mathcal{S}}$. The tokens ($\mathbf{T}^{\mathcal{T}}, \mathbf{T}^{\mathcal{S}}$) are generated through tokenization, either at the image-level or conducted at the feature-level by employing patching techniques.

\subsection{Vanilla Knowledge Distillation}\label{Distillation}
We begin with an overview of vanilla knowledge distillation for a comprehensive understanding. 
The vanilla knowledge distillation~\cite{hinton2015distilling} is with 
the objective of transferring knowledge from a teacher model to a student model by utilizing soft outputs: 
\begin{equation}
    p(\boldsymbol{x}, \tau) = \frac{\textrm{exp}(\boldsymbol{z}_i(\boldsymbol{x})/\tau)}{\sum_j\textrm{exp}(\boldsymbol{z}_j(\boldsymbol{x})/\tau)}, 
\label{soft labels}
\end{equation}
where ${\boldsymbol{z}(\boldsymbol{x})}=[\boldsymbol{z}_1(\boldsymbol{x}),\cdots,\boldsymbol{z}_K(\boldsymbol{x})]$ is the logit vector produced by the model, softened by the temperature hyperparameter $\tau$. 
A higher value of $\tau$ leads to a softer probability distribution across the classes. We optimize the student model by incorporating both ground-truth labels and soft targets generated by the pre-trained teacher. Hence the student network is trained to minimize the Kullback-Leibler (KL) divergence between the soft targets ${p}^{\mathcal{T}}$ from the teacher network and the predicted probabilities ${p}^{\mathcal{S}}$ from the student network:
\begin{equation}
    \mathcal{L}_{KD}({p}^{\mathcal{S}},{p}^{\mathcal{T}})=\frac{1}{N} \sum_{i=1}^{N} \textrm{KL}\left({p}^{\mathcal{S}}, {p}^{\mathcal{T}}\right).
\label{equa:kd}
\end{equation}
The student network is further trained using the classical Cross-Entropy (CE) loss with the hard labels. The total loss can be formulated as follows:
\begin{equation}
    \mathcal{L}_\textrm{logit}=\mathcal{L}_{CE}({p}^{s},{y})+\lambda \mathcal{L}_{KD}({p}^{\mathcal{S}},{p}^{\mathcal{T}}),
\label{logit distillation loss}
\end{equation}
where $\lambda$ is the trade-off weight. 
And $\mathcal{L}_{CE}$ represents the Cross-Entropy (CE) loss between the hard labels and the predictions. The subscript ``logit'' in $\mathcal{L}_\textrm{logit}$ indicates that vanilla knowledge distillation is performed by utilizing the final logits.

\label{03Pre}
\section{Methodology}

\begin{figure*}

\centering 
\includegraphics[width=0.9\textwidth]{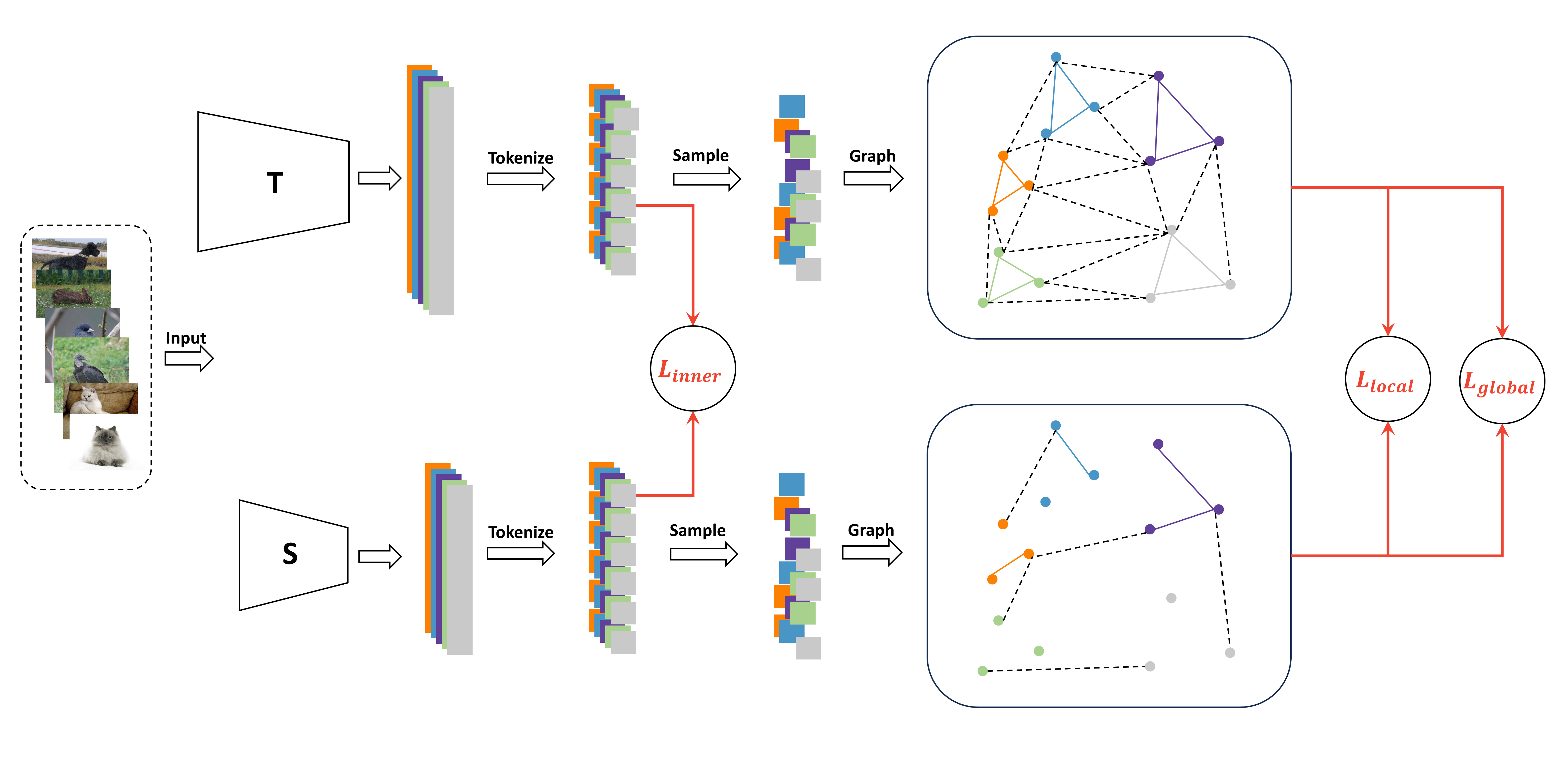}
\caption{
Illustration of the overall framework of our proposed \name. We distill the token-level structural knowledge from the teacher to the student by constructing a token-wise graph. Besides, we also consider the $\mathcal{L}_\textrm{inner}$ loss to transmit the contextual semantics within the image.}
\label{fig:framework} 
\end{figure*}

This section  presents our proposed method that involves the construction of a graph to capture token-level relational knowledge. We begin by introducing the token patching technique, which aims to create a token-level graph for the distillation process. To 
reduce the computational complexity associated with constructing large-scale graphs, we introduce a random sampling strategy that selects tokens from different images. This allows us to effectively represent the token-level relationships using a more manageable graph structure. With the  k-Nearest Neighbors (\textit{k}-NN) algorithm, we may construct the token-level graph to distill the token-wise relational knowledge from the teacher to the student. Moreover, we employ the inner-instance contextual loss to transfer the image-level semantic knowledge. Finally, we outline the complete algorithm of our training structure and introduce the dynamic temperature mechanism for better optimization. The overall framework of our method is illustrated in Fig.~\ref{fig:framework}.

\subsection{Token-wise Graph Distillation}
\subsubsection{Token Patching} 
As described in 
\cite{dosovitskiy2020image}, the ViT-like architectures perform token patching at the image level. For an input image $\boldsymbol{x}\in \mathbb{R}^{C \times H\times W}$, token patching reshapes the image into $M$ flattened patches $\boldsymbol{x_p}\in \mathbb{R}^{M\times D}$. Here, $M = HW/P^2$ (patch size $P$) represents the number of patches, and the dimension $D$ is calculated as $D = P^2 C$. Given a batch of data with batch size $B$, we feed the tokenized images into both teacher and student networks to obtain the token feature $\mathbf{T}^{\mathcal{T}}$ and $\mathbf{T}^{\mathcal{S}}$, respectively. In contrast, in CNN-like architectures, we adopt a slightly different approach for token patching, operating at the feature level. The feature map of the $\ell$-th intermediate layer can be denoted as $F^\ell \in \mathbb{R}^{C_\ell \times H_\ell \times W_\ell}$, where $C_\ell$, $H_\ell$, and $W_\ell$ represent the channel number, height, and width of the feature map in the $\ell$-th layer, respectively. For simplicity, we choose the feature map from the penultimate layer. Since the teacher and student networks may have different sizes, the shapes of feature maps can also differ. 
Hence, we employ different patch sizes for the teacher's feature and the student's feature. This ensures that an equal number of patches is selected, thereby maintaining the same number of nodes in the constructed graph throughout the distillation process.

\subsubsection{Random Token Sampling}
We construct graphs at the token level for both the teacher and student networks. However, considering all tokens in a batch may result in an excessively large-scale graph, especially in the case of datasets like ImageNet~\cite{russakovsky2015imagenet}, where a graph with $B\times16\times16$ nodes would need to be built in a batch of size $B$. This poses optimization challenges during the training process. To address this issue, we adopt a sampling strategy to randomly select tokens from different images. This approach allows us to create a relatively smaller graph while retaining the crucial relationship knowledge among tokens across images. Specifically, we generate a random sampling denoted as $\mathcal{RS}$ to select tokens from instances generated by the student network. This sampling is evenly distributed among all instances in the batch. To ensure correspondence between the selected tokens from the student and teacher networks, we apply the same sampling strategy $\mathcal{RS}$ to select the corresponding tokens from the teacher's tokenized features.

\subsubsection{k-NN Graph Construction}
For a pair of tokens, we consider them to be semantically related if their embeddings exhibit a small distance inbetween. 
We thereby construct a $k$-nearest neighbor (\textit{k}-NN) graph, where each vertex is connected to its $k$ closest neighbors based on the minimum distance. By using this \textit{k}-NN graph construction, we filter out the least semantically related token pairs. This approach is preferred over introducing a fully connected graph, as randomly sampled tokens typically have only a few related samples, which are sufficient to capture the relationship information without introducing redundant information. We choose to use the symmetric \textit{k}-NN graph construction for convenience, and our adjacent matrix has the form of:
\begin{equation}
    \label{eq:knn}
    \mathbf{A}(\mathbf{T}_i,\mathbf{T}_j) =
    \begin{cases}
        0, & \mathbf{T}_i \notin \textit{k}\textrm{-NN}(\mathbf{T}_j), \\
        e^{ - \frac{1}{2\sigma} \lVert \mathbf{T}_i - \mathbf{T}_j \rVert^{2} }, & \text{otherwise}.
    \end{cases}
\end{equation}
Thus, we construct the attributed graphs for the teacher network and the student network, denoted as $\mathbf{G}^{\mathcal{T}} = \{{\mathbf{A}}^{\mathcal{T}},{\mathbf{T}}^{\mathcal{T}}\}$ and $\mathbf{G}^\mathcal{S} = \{{\mathbf{A}^{\mathcal{S}}},{\mathbf{T}}^{\mathcal{S}}\}$, respectively. Here, $\mathbf{A}^{\{\mathcal{T},\mathcal{S}\}}$ and $\mathbf{T}^{\{\mathcal{T},\mathcal{S}\}}$ represent the adjacency matrices and node attributes of the attributed graphs for teacher $\mathcal{T}$ and student $\mathcal{S}$. It is worth noting that graph $\mathbf{G}^{\mathcal{T}}$ remains fixed as the teacher network has already been well optimized. On the other hand,  graph $\mathbf{G}^{\mathcal{S}}$ is updated during training, involving updates to both the node attributes and the graph topology, as the student network learns and adapts to the distillation process.

\subsubsection{Local Preserving Loss}
As mentioned 
above, the selected tokens with similar content should be located in the neighborhood of each other. To preserve the local structure, we adopt the concept of local preserving loss~\cite{yang2020distilling}. Specifically, we utilize the softmax probability distribution of the similarities among $\mathbf{T}_i$ and its neighboring tokens $\mathbf{T}_j$ to capture the local structure. The local preserving loss $\mathcal{L}_\textrm{local}$ aims to distill the student network to mimic the local structure observed in the teacher's embedding space. This is achieved by minimizing the KL divergence between the softmax probability distributions of the student and teacher embeddings:
\begin{equation}
    \label{eq:lsp}
    \mathcal{L}_{\textrm{local}} = \sum_{i \in \mathbf{T}} {\textrm{KL}} \Big( \underset{j \in \mathbf{T}}{\textrm{softmax}} \big(\mathbf{A}_{ij}^\mathcal{S} \big) \, \Vert \, \underset{j \in \mathbf{T}}{\textrm{softmax}} \big( \mathbf{A}_{ij}^\mathcal{T} \big) \Big).
\end{equation}

\subsubsection{Global Relationship Loss}
Only relying on the local preserving loss may result in a loss of global topology of the teacher's embedding. Therefore, our objective 
also considers  
transferring the global topology of teacher tokens to the student network. To achieve this, we draw inspiration from the concept of contrastive knowledge distillation~\cite{tian2019contrastive} and extend it to the token-graph version. Specifically, we aim to maximize the similarity between pairs of teacher and student tokens that share the corresponding node (referred to as positive pairs), meanwhile minimizing the similarity with unmatched token nodes (negative pairs). This contrastive learning approach helps the student network capture the underlying relationships and structure present in the teacher's token graph, enabling a more comprehensive knowledge transfer. Considering that the dimensionality of the student and teacher tokens may differ, we introduce a linear projection, denoted as $\textrm{Proj}$, for feature alignment. Consequently, the similarity between tokens from the teacher and student can be computed as follows:
\begin{equation}
    \mathcal{SIM}(\mathbf{T}^\mathcal{S},\mathbf{T}^\mathcal{T}) = \frac{\textrm{Proj
    }(\mathbf{T}^\mathcal{S})'\cdot{\mathbf{T}^\mathcal{T}}}{\lVert\textrm{Proj
    }(\mathbf{T}^\mathcal{S})\rVert\cdot\lVert{\mathbf{T}^\mathcal{T}
}\rVert}.
\label{eq:similar equation}
\end{equation}
With the calculated similarity $\mathcal{SIM}$, we conduct graph-based contrastive distillation with adapting the InfoNCE loss \cite{oord2018representation}:
\begin{equation}
    \label{eq:gcrd}
    \mathcal{L}_{\textrm{global}} = - \sum_{i \in \mathbf{T}} \textrm{ log} \frac{\textrm{ exp} \big( \frac{\mathcal{SIM} ( \mathbf{T}_i^\mathcal{S} , \mathbf{T}_i^\mathcal{T} )}{\tau_g} \big)}{\sum_{j \in \mathbf{T}}\textrm{ exp} \big( \frac{\mathcal{SIM} ( \mathbf{T}_i^\mathcal{S} , \mathbf{T}_j^\mathcal{T})}{\tau_g} \big)},
\end{equation}
where $\tau_g$ is the temperature scaling hyper-parameter. 


\subsubsection{Dynamic Temperature Adjustment}
During the initial stages of training, the tokens of student model show limited semantic understanding, thus we may separate the positive pairs and negative samples using Eq.~\ref{eq:gcrd} easily, leading a sketchy global topology of graph. However, after the student model is trained to a certain extent, the graph of student nodes may indeed capture the coarse topology of the teacher but might lack the fine-grained structural details. Consequently, it becomes challenging to effectively differentiate positive pairs from hard negative samples, thereby impeding further improvement of the student model's performance. 

To mitigate this issue, we propose a temperature strategy that regulates the global contrastive loss by using a dynamic graph temperature strategy.  Our temperature strategy draws inspiration from the work of Wang \etal \cite{wang2020understanding}, that investigates the role of temperature in contrastive learning. They highlight the significance of the temperature coefficient and its impact on the learning process. Specifically, they observe that a lower temperature tends to prioritize hard negative samples, resulting in a more dispersed local structure and a potentially more uniform distribution of embeddings. Drawing from the insights of Wang \etal \cite{wang2020understanding}, we employ a dynamic temperature strategy. Specifically, during the initial stage of graph distillation, we use a relatively high temperature to facilitate the transfer of coarse topological information from  teacher to student. During the stage of training, we gradually decrease the temperature $\tau_g$ to encourage the student to learn a more fine-grained representation of the teacher's topology. For simplicity, the form of our dynamic temperature $\tau_g$ is:
\begin{equation}
    \label{eq:dynamic temperature}
    \tau_g =
    \begin{cases}
        \tau_g^\textrm{init}, & \textit{epoch} \leq \textrm{W}_\textrm{U}, \\
        \tau_g^\textrm{init}/\log_{\textrm{W}_\textrm{U}}(\textit{epoch}), & \textit{epoch} > \textrm{W}_\textrm{U}.
    \end{cases}
\end{equation}
Here $\tau_g^\textrm{init}$ is the initial temperature, $\textit{epoch}$ denotes the epoch in training process, and $\textrm{W}_\textrm{U}$ indicates the number of warm-up epochs.

\subsection{Contextual Loss}
Except for the cross-image relationship, we also incorporate contextual similarity loss within each individual image to preserve contextual information during the distillation process. To capture the contextual semantics, we utilize the penultimate feature maps. Specifically, for the $i$-th image, the feature maps in the penultimate layer are represented as $\mathbf{F}^\mathcal{T}_i \in \mathbb{R}^{N\times D^\mathcal{T}}$ and $\mathbf{F}^\mathcal{S}_i \in \mathbb{R}^{N\times D^\mathcal{S}}$, corresponding to the teacher and student networks, respectively. The contextual similarity measures the token relationship within each individual instance and can be expressed as follows:
\begin{equation}
    \mathrm{CS} = \textrm{Softmax}(\frac{\mathbf{F}\cdot \mathbf{F}'}{\sqrt{D}}). 
    \label{eq:contexture similarity}
\end{equation}
The contextual similarity $\mathrm{CS}$ measures the contextual similarity of the feature generated by the student and the teacher, respectively. We argue that a good student should embody similar contextual similarity as the teacher. To quantify the discrepancy in contextual similarity, we utilize the mean squared error (MSE) loss. Therefore, our inner-instance contextual similarity loss can be formulated as follows:
\begin{equation}
    \mathcal{L}_\textrm{inner} = \mathrm{MSE}(\mathrm{CS}^\mathcal{T}, \mathrm{CS}^\mathcal{S}).
    \label{inner loss}
\end{equation}
The contextual similarity loss plays a crucial role in preserving contextual semantics in knowledge distillation. 

In summary, our total loss has the form of:
\begin{equation}
        \mathcal{L}_\textrm{total} = \mathcal{L}_\textrm{logit} + \alpha\mathcal{L}_\textrm{inner} + \beta\mathcal{L}_\textrm{local} + \gamma\mathcal{L}_\textrm{global},
\label{eq:total loss}
\end{equation}
where $\mathcal{L}_\textrm{logit}$ is the logit distillation loss defined in Eq.~\ref{logit distillation loss}, and $\alpha, \beta, \gamma$ are the trade-off coefficients.


\begin{algorithm}[t] 
   \caption{The \name Algorithm} 
   \label{alg1} 
   \begin{algorithmic}[1] 
   \renewcommand{\algorithmicrequire}{ \textbf{Input}}
   \REQUIRE:
   
   $\mathcal{D}$ \textcolor[RGB]{77,138,138}{ \# set of images and the corresponding labels}
    
    $\theta$    \textcolor[RGB]{77,138,138}{ \# the parameters  of the model} 
    
    $\tau,\tau_g$    \textcolor[RGB]{77,138,138}{ \# the temperature for soft logit distillation, token-wise graph global loss}

   \renewcommand{\algorithmicrequire}{ \textbf{Output}}
   \REQUIRE: A well-trained student model      
   \WHILE{$\theta$ is not converged}
   \STATE Sample ($\boldsymbol{x}$,$y$) from the training dataset $\mathcal{D}$
   \STATE Generate feature representation and the prediction probabilities $\mathbf{F}^\mathcal{T}, \mathbf{F}^\mathcal{S}$, $\boldsymbol{z}^\mathcal{T},\boldsymbol{z}^\mathcal{S}$ using the teacher $\mathcal{T}$ and the student $\mathcal{S}$.
   \STATE Calculate the logit distillation loss using Eq.~\eqref{logit distillation loss}.
   \STATE Same token sampling strategy over the feature representation $\mathbf{F}^
   \mathcal{T},\mathbf{F}^\mathcal{S}$ across instances in a batch dataloader.
   \STATE Generate fixed token graph $\mathbf{G}^\mathcal{T}$, and learning graph $\mathbf{G}^\mathcal{S}$ using \textit{k}-NN graph construction with the adjacent matrix Eq.~\eqref{eq:knn}.
   \STATE Conduct dynamic graph temperature $\tau_g$ over the training epochs by using Eq.~\eqref{eq:dynamic temperature}.
    \STATE Compute the graph local preserving loss Eq.~\eqref{eq:lsp} and global graph InfoNCE loss Eq.~\eqref{eq:gcrd} to preserve the topology of selected tokens.
    \STATE Compute contextual loss by using Eq.~\eqref{inner loss} to transfer the contextual semantics of the teacher.
   \STATE Update $\theta$ by compute gradient by using the total loss in Eq.~\eqref{eq:total loss}. 
   \ENDWHILE
   \end{algorithmic}
\end{algorithm}

\label{04Method}

\section{Experiments}
To validate the effectiveness of our proposed method, we perform extensive evaluations by comparing \name with several state-of-the-art approaches. Our evaluations encompass diverse image classification tasks, including the challenging scenario of imbalanced dataset classification. We consider two distinct network architectures: Convolutional Neural Networks (CNNs) and Vision Transformers (ViTs), which represent prominent models in the field of computer vision. By conducting this comprehensive comparative analysis, we aim to assess the generalization and effectiveness of our approach across various model architectures and classification scenarios.

\subsection{Main Results}
\subsubsection{Basic Settings}
{\bf Baselines.} \quad 
We compare our approach with two kinds of prevalent and advanced knowledge distillation baselines, \ie, individual distillation and relation-based (including graph-based) distillation: 
\begin{itemize}
\setlength{\itemsep}{0pt}
\setlength{\parsep}{0pt}
\setlength{\parskip}{0pt}
	\item \textbf{Individual distillation} includes the vanilla KD~\cite{hinton2015distilling}, FitNet~\cite{romero2014fitnets}, AT~\cite{zagoruyko2016paying}, SP~\cite{tung2019similarity}, AB~\cite{heo2019knowledge}, DIST~\cite{huang2022knowledge},
 SemCKD~\cite{chen2021cross}, 
	DKD~\cite{zhao2022decoupled}.
	\item \textbf{Relation-based distillation} includes  SP~\cite{tung2019similarity}, RKD~\cite{park2019relational},CRD~\cite{tian2019contrastive}, CC~\cite{peng2019correlation} and two graph-based distillations, \ie, IRG~\cite{liu2019IRG} and HKD~\cite{zhou2021distilling}.
\end{itemize}

{\noindent \bf Datasets.} \quad We assess the effectiveness of \name on two widely used datasets: CIFAR-100~\cite{krizhevsky2009learning} and ImageNet~\cite{russakovsky2015imagenet}. The CIFAR-100 dataset consists of color-scaled images with dimensions $32\times32$, featuring objects from 100 different classes. It is divided into a training set of 50,000 images and a validation set of 10,000 images. Additionally, we also conduct experiments on long-tailed CIFAR-100 (CIFAR-100-LT) , a modified version of CIFAR-100 where the number of instances in the minor classes is reduced. For large-scale tasks, we utilize the ImageNet dataset, where all the images are resized to dimensions ($224\times224$). Similar to CIFAR-100-LT, we evaluate the classification performance on ImageNet-LT~\cite{liu2019large} (the long-tailed version of ImageNet-1K with imbalance rate of $10$), which serves as an experimental benchmark for testing on imbalanced datasets. We operate the standard data augmentation (\ie, flipping and random cropping) and normalization as conducted in \cite{he2016deep,zagoruyko2016wide,huang2017densely} on the above two datasets.
\\

{\noindent \bf Networks.} \quad We conduct a series of classification with typical mainstream architectures and lightweight networks on the CIFAR-100 and ImageNet datasets. Our implementation for CIFAR-100 follows the practice in~\cite{chen2021cross}. A variety of teacher-student pairs based on popular visual network architectures are tested, including ResNet~\cite{he2016deep}, VGG~\cite{simonyan2014very}, WideResNet (WRN)~\cite{zagoruyko2016wide}, and several lightweight networks such as MobileNet~\cite{howard2017mobilenets} and ShuffleNet~\cite{zhang2018shufflenet}. 
In CIFAR-100 training, we adapt ResNet and WideResNet to small-scale datasets by using PreAct layer~\cite{he2016deep}. 
We modify the first convolutional layer of ResNet to the kernel size of $3 \times 3$, strides of 1, and padding of 1. 
For the ImageNet evaluation, we further investigate the performance of our approach using Vision Transformer (ViT)-based networks. Following the architectural guidelines established by DeiT \cite{touvron2021training}, we conduct experiments using DeiT models and assess their performance when combined with various existing KD methods.
\\

\noindent {\bf Training Details for CIFAR-100/CIFAR-100-LT.} \quad 
For the training of CIFAR-100, we adopt SGD optimizier with 0.9 Nesterov momentum, the total training epoch is predetermined to 240, the warm-up epoch is set to 15, and we divide the learning rate by 10 at epochs 150, 180, and 210. The initial learning rate is set to 0.01 for lightweight architectures, and 0.05 for the other series. We train with weight decay $5e^{-4}$ for regularization. We set the temperature $\tau=4$ for conventional KD loss and the initial temperature $\tau_g=0.1$ for graph-based contrastive loss.
\\

\noindent {\bf Training Details for ImageNet/ImageNet-LT.}\quad For training on the ImageNet dataset, we adopt the recommended practices provided by the PyTorch official documentation. Our models are trained by a total of 100 epochs for CNN-like networks and 200 epochs for ViT-based architectures (both use 10-epoch warm up). In CNN-based training, we initialize the learning rate to 0.1 and decrease it by a factor of 10 every 30 epochs. The batch size is set to $128 \times 4$, and the weight decay rate is set to $1e^{-4}$. Additionally, we utilize a cosine schedule with a 10-epoch warm-up phase during training. 
In ViT training, we use the AdamW~\cite{loshchilov2017decoupled} optimizer with the learning rate of $5e^{-4}$. The learning rate is adjusted with the cosine annealing strategy. Unless otherwise stated, all experiments on the ImageNet dataset are conducted using 4 GPUs of RTX3090. The optimal results are determined by maximizing the top-1 accuracy on the validation set.
\\

\begin{table*}[htbp]
	\centering
    \caption{Top-1 test accuracy (\%) of various distillation approaches on CIFAR-100. The teacher and student pairs share similar architectures. Each experiment is repeated three times, and we report the mean and standard deviation of the top-1 accuracy. The best results appear in \textbf{\color{blue} blue bold}.
	}
	\resizebox{0.90\textwidth}{!}{
		\begin{tabular}{l|l|cccccc}
			\toprule
			\multirow{2}{*}{Type} & \multirow{2}{*}{Student} & ResNet-8$\times$4 & VGG-8 & ResNet20 & WRN-40-1 & WRN-16-2 & ResNet32 \\
			& & 72.51 $\pm$ 0.29 & 70.46 $\pm$ 0.29 & 69.06 $\pm$ 0.22 & 71.98 $\pm$ 0.17 & 73.43 $\pm$ 0.22 & 71.14 $\pm$ 0.25  \\
			\midrule
			\multirow{2}{*}{---} &\multirow{2}{*}{Teacher} & ResNet-32$\times$4 & VGG-13 & ResNet56 & WRN-40-2 & WRN-40-2 & ResNet110  \\
			& & 79.42 & 74.64 & 73.44 & 76.31 & 76.31 & 74.31 \\
			\midrule
			\multirow{8}{*}{Individual} & KD \cite{hinton2015distilling} & 74.12 $\pm$ 0.15 & 72.66 $\pm$ 0.13 & 70.66 $\pm$ 0.22 & 73.42 $\pm$ 0.22 & 74.92 $\pm$ 0.20 & 73.02 $\pm$ 0.16 \\
            & FitNet \cite{romero2014fitnets} & 73.89 $\pm$ 0.22 & 73.54 $\pm$ 0.12 & 71.52 $\pm$ 0.16 & 74.12 $\pm$ 0.20 & 75.75 $\pm$ 0.12 & 72.52 $\pm$ 0.07 \\
			& AT \cite{zagoruyko2016paying} & 74.57 $\pm$ 0.17 & 73.63 $\pm$ 0.12 & 71.76 $\pm$ 0.14 & 74.43 $\pm$ 0.11 & 75.28 $\pm$ 0.13 & 73.32 $\pm$ 0.11 \\
			& FT \cite{kim2018paraphrasing} & 74.26 $\pm$ 0.21 & 73.44 $\pm$ 0.21 & 71.68 $\pm$ 0.13 & 74.17 $\pm$ 0.19 & 75.32 $\pm$ 0.22 & 73.57 $\pm$ 0.21 \\
			& AB \cite{heo2019knowledge} & 74.39 $\pm$ 0.11 & 73.68 $\pm$ 0.21 & 72.00 $\pm$ 0.25 & 74.18 $\pm$ 0.22 & 74.88 $\pm$ 0.13 & 74.11 $\pm$ 0.11 \\
			& DIST \cite{huang2022knowledge} & 76.11 $\pm$ 0.23 & 74.41 $\pm$ 0.12 & 71.88 $\pm$ 0.09 & 76.22 $\pm$ 0.24 & 76.32 $\pm$ 0.17 & 73.88 $\pm$ 0.16 \\
			& SemCKD \cite{chen2021cross} & {75.58 $\pm$ 0.22} & {74.42 $\pm$ 0.21} & {71.98 $\pm$ 0.17} & {74.78 $\pm$ 0.21} & {75.42 $\pm$ 0.15} & {74.12 $\pm$ 0.22} \\
            & DKD \cite{zhao2022decoupled} & 76.32 $\pm$ 0.26 & 74.68 $\pm$ 0.23 & 71.79 $\pm$ 0.17 & 76.11 $\pm$ 0.17 & {\bf \color{blue}76.55 $\pm$ 0.14} & 74.11 $\pm$ 0.17 \\
			\midrule
			\multirow{4}{*}{Relation-based} & SP \cite{tung2019similarity} & 73.90 $\pm$ 0.17 & 73.44 $\pm$ 0.21 & 71.48 $\pm$ 0.11 & 73.17 $\pm$ 0.21 & 75.34 $\pm$ 0.21 & 73.63 $\pm$ 0.21 \\
            & CRD \cite{tian2019contrastive} & 75.59 $\pm$ 0.23 & 73.88 $\pm$ 0.18 & 71.68 $\pm$ 0.11 & 75.51 $\pm$ 0.22 & 76.01 $\pm$ 0.11 & 73.48 $\pm$ 0.16 \\
            & RKD \cite{park2019relational} & 75.11 $\pm$ 0.13 & 73.62 $\pm$ 0.12 & 71.32 $\pm$ 0.13 & 75.22 $\pm$ 0.20 & 75.88 $\pm$ 0.13 & 73.58 $\pm$ 0.12 \\
            & CC \cite{peng2019correlation} & 75.09 $\pm$ 0.21 & 73.88 $\pm$ 0.16 & 71.28 $\pm$ 0.14 & 75.31 $\pm$ 0.22 & 75.66 $\pm$ 0.13 & 73.48 $\pm$ 0.11 \\
            \midrule
            \multirow{3}{*}{Graph-based}& 
             IRG \cite{liu2019IRG} & 74.79 $\pm$ 0.21 & 73.68 $\pm$ 0.14 & 71.38 $\pm$ 0.14 & 75.21 $\pm$ 0.24 & 75.58 $\pm$ 0.12 & 73.41 $\pm$ 0.16 \\
             & HKD \cite{zhou2021distilling} & 76.21 $\pm$ 0.21 & 74.21 $\pm$ 0.11 & 72.11 $\pm$ 0.10 & 75.99 $\pm$ 0.22 & 76.21 $\pm$ 0.10 & 74.42 $\pm$ 0.14 \\
            & \name (Ours) & \textbf{\color{blue}76.42 $\pm$ 0.22} & \textbf{\color{blue}74.89 $\pm$ 0.12} & \textbf{\color{blue}72.21 $\pm$ 0.23} & \textbf{\color{blue}76.12 $\pm$ 0.21} & {76.22 $\pm$ 0.17} & \textbf{\color{blue}74.43 $\pm$ 0.28} \\
			\bottomrule
		\end{tabular}
	}
	\vspace{3pt}
	\label{Tbl:CIFAR-100-1}
\end{table*} 

\vspace{3pt}

\begin{table*}[t]
	\centering
    \caption{Top-1 test accuracy (\%) of the various distillation approaches with heterogeneous  teacher-student pairs on CIFAR-100. Each experiment is repeated three times, and we report the mean and standard deviation of the top-1 accuracy.  
    The best results appear in \textbf{\color{blue}blue bold}. }
	\resizebox{0.90\textwidth}{!}{
		\begin{tabular}{l|l|cccccc}
			\toprule
			\multirow{2}{*}{Type} & \multirow{2}{*}{Student} & ShuffleV1 & WRN-16-2 & VGG-8 & MobileV2 & MobileV2 & ShuffleV1  \\
			& & 70.50 $\pm$ 0.22 & 73.43 $\pm$ 0.22 & 70.46 $\pm$ 0.29 & 64.60 $\pm$ 0.32 & 64.60 $\pm$ 0.32 & 70.50 $\pm$ 0.22   \\
			\midrule
			\multirow{2}{*}{---} &\multirow{2}{*}{Teacher} & ResNet-32x4 & ResNet-32x4 & ResNet50 & WRN-40-2 & VGG-13 & WRN-40-2  \\
			& & 79.42 & 79.42 & 79.10 & 76.31 & 74.64 & 76.31  \\
			\midrule
			\multirow{8}{*}{Individual} & KD \cite{hinton2015distilling} & 74.00 $\pm$ 0.16 & 74.90 $\pm$ 0.29 & 73.81 $\pm$ 0.24 & 69.07 $\pm$ 0.26 & 67.37 $\pm$ 0.22 & 74.83 $\pm$ 0.13   \\
            & FitNet \cite{romero2014fitnets} & 74.82 $\pm$ 0.13 & 74.70 $\pm$ 0.35 & 73.72 $\pm$ 0.18 & 68.71 $\pm$ 0.21 & 63.16 $\pm$ 0.23 & 74.11 $\pm$ 0.23   \\
			& AT \cite{zagoruyko2016paying} & 74.76 $\pm$ 0.19  & 75.38 $\pm$ 0.18 & 73.45 $\pm$ 0.17 & 68.64 $\pm$ 0.12 & 63.42 $\pm$ 0.21 & 73.73 $\pm$ 0.19  \\
			& FT \cite{kim2018paraphrasing} & 74.28 $\pm$ 0.12 & 74.85 $\pm$ 0.35 & 73.75 $\pm$ 0.21 & 68.91 $\pm$ 0.21 & 65.70 $\pm$ 0.28 & 74.41 $\pm$ 0.26   \\
			& AB \cite{heo2019knowledge} & 74.22 $\pm$ 0.29 & 75.22 $\pm$ 0.07 & 73.61 $\pm$ 0.21 & 68.68 $\pm$ 0.24 & 66.21 $\pm$ 0.20 & 74.66 $\pm$ 0.21   \\
            & DIST \cite{huang2022knowledge} & 76.11 $\pm$ 0.17 & 75.95 $\pm$ 0.07 & 75.22 $\pm$ 0.16 & 69.45 $\pm$ 0.21 & 69.23 $\pm$ 0.20 & 75.89 $\pm$ 0.22   \\
			& SemCKD \cite{chen2021cross} & 75.41 $\pm$ 0.11 & 75.65 $\pm$ 0.23 & 74.68 $\pm$ 0.22 & 69.88 $\pm$ 0.30 & 68.78 $\pm$ 0.22 & 74.81 $\pm$ 0.21  \\
            & DKD \cite{zhao2022decoupled} & {\bf \color{blue}76.42 $\pm$ 0.11} & 76.11 $\pm$ 0.22 & 75.44 $\pm$ 0.22 & 69.47 $\pm$ 0.21 & 69.71 $\pm$ 0.26 & {\bf \color{blue}76.11 $\pm$ 0.13}   \\
			\midrule
            \multirow{4}{*}{Relation-based} & SP \cite{tung2019similarity} & 73.80 $\pm$ 0.21 & 75.16 $\pm$ 0.32 & 73.86 $\pm$ 0.15 & 68.48 $\pm$ 0.22 & 65.42 $\pm$ 0.21 & 74.01 $\pm$ 0.11   \\
            & CRD \cite{tian2019contrastive} & 75.46 $\pm$ 0.23 & 75.70 $\pm$ 0.29 & 74.42 $\pm$ 0.21 & 69.87 $\pm$ 0.17 & 69.73 $\pm$ 0.21 & 76.05 $\pm$ 0.23   \\
            & RKD \cite{park2019relational} & 74.46 $\pm$ 0.11 & 75.20 $\pm$ 0.26 & 74.12 $\pm$ 0.11 & 68.81 $\pm$ 0.27 & 66.23 $\pm$ 0.24 & 75.05 $\pm$ 0.21   \\
            & CC \cite{peng2019correlation} & 74.69 $\pm$ 0.13 & 75.40 $\pm$ 0.21 & 74.15 $\pm$ 0.26 & 68.98 $\pm$ 0.15 & 66.73 $\pm$ 0.24 & 75.28 $\pm$ 0.17   \\
            \midrule
			\multirow{3}{*}{Graph-based} & IRG \cite{liu2019IRG} & 74.82 $\pm$ 0.24 & 74.11 $\pm$ 0.04 & 74.77 $\pm$ 0.25 & 68.38 $\pm$ 0.17 & 68.42 $\pm$ 0.11 & 74.99 $\pm$ 0.11 \\
             & HKD \cite{zhou2021distilling} & 75.99 $\pm$ 0.23 & 75.61 $\pm$ 0.14 & 74.42 $\pm$ 0.16 & 69.99 $\pm$ 0.14 & 69.21 $\pm$ 0.14 & 75.89 $\pm$ 0.14 \\
             & \name (Ours) & \textbf{\color{blue}76.42 $\pm$ 0.22} & \textbf{\color{blue}76.55 $\pm$ 0.11} & \textbf{\color{blue}75.98 $\pm$ 0.24} & \textbf{\color{blue}70.11 $\pm$ 0.14} & \textbf{\color{blue}70.32 $\pm$ 0.21} & {76.08 $\pm$ 0.11} \\
			\bottomrule
		\end{tabular}
	}
	\vspace{3pt}
	\label{Tbl:CIFAR-100-2}
\end{table*}

\begin{table*}[t]
\setlength{\tabcolsep}{4.5pt}
\begin{center}
\caption{
Top-1 and Top-5 accuracies (\%) of student network ResNet-18 on the ImageNet and ImageNet-LT validation set. We use ResNet-34 released by PyTorch official as the teacher network, and follow the standard training practice of ImageNet on PyTorch guideline. The best results appear in {\bf \color{blue}blue bold}, and the downarrow \textcolor[RGB]{77,138,138}{$\downarrow$} denotes the accuracy decrease  comparing to the standard dataset.
}
\begin{tabular}{cc|ll|llllll|l}
\toprule
 DataSet & Metric & Teacher & Student & KD\cite{hinton2015distilling} & AT\cite{zagoruyko2016paying} &  RKD\cite{park2019relational} &   
 DKD\cite{zhao2022decoupled} & 
 IRG\cite{liu2019IRG} &
 HKD\cite{zhou2021distilling} & \name \\
\midrule
\multirow{2}{*}{ImageNet} & Top-1 & 73.31 & 69.75 & 70.70 & 70.66 & 70.88 & 71.17 & 70.66 & 71.21 & \textbf{\color{blue}71.31} \\
& Top-5 & 91.42 & 89.07 & 90.00 & 89.88 & 89.99 &  90.19 & 89.64 & 90.18 & \textbf{\color{blue}90.41}\\
\midrule
\multirow{2}{*}{ImageNet-LT} & Top-1 & 50.11\textcolor[RGB]{77,138,138}{\tiny ($\downarrow$23.20)} & 43.22\textcolor[RGB]{77,138,138}{\tiny ($\downarrow$26.53)} & 46.70\textcolor[RGB]{77,138,138}{\tiny ($\downarrow$24.00)} & 46.88\textcolor[RGB]{77,138,138}{\tiny ($\downarrow$23.78)}  & 48.31\textcolor[RGB]{77,138,138}{\tiny ($\downarrow$22.57)} &  48.32\textcolor[RGB]{77,138,138}{\tiny ($\downarrow$22.85)} & 46.31\textcolor[RGB]{77,138,138}{\tiny ($\downarrow$24.35)} & 49.21\textcolor[RGB]{77,138,138}{\tiny ($\downarrow$22.00)} & \textbf{\color{blue}50.32}\textcolor[RGB]{77,138,138}{\tiny ($\downarrow$20.99)} \\
& Top-5 & 77.62\textcolor[RGB]{77,138,138}{\tiny ($\downarrow$13.80)} & 75.42\textcolor[RGB]{77,138,138}{\tiny ($\downarrow$13.65)} & 76.11\textcolor[RGB]{77,138,138}{\tiny ($\downarrow$13.89)} & 76.43\textcolor[RGB]{77,138,138}{\tiny ($\downarrow$13.45)}  & 76.62\textcolor[RGB]{77,138,138}{\tiny ($\downarrow$13.37)}  & 77.03\textcolor[RGB]{77,138,138}{\tiny ($\downarrow$13.16)} & 76.04\textcolor[RGB]{77,138,138}{\tiny ($\downarrow$13.60)} & 77.18\textcolor[RGB]{77,138,138}{\tiny ($\downarrow$13.00)} & \textbf{\color{blue}77.88}\textcolor[RGB]{77,138,138}{\tiny ($\downarrow$12.53)}\\
\bottomrule
\end{tabular}
\label{Tbl:ImageNet}
\end{center}
\vspace{-5pt}
\end{table*}


\subsubsection{Results with CNN-based Architectures}
{\bf Results on CIFAR-100.}\quad
In Tables~\ref{Tbl:CIFAR-100-1} and ~\ref{Tbl:CIFAR-100-2}, we present a comprehensive comparison of our proposed \name  with several established distillation techniques on the CIFAR-100 dataset. We consider two scenarios: teacher-student pairs with similar architectures (ResNet110/ResNet-32, VGG-13/VGG-8) and teacher-student pairs with heterogeneous architectures (ResNet-32$\times$4/ShuffleV1, VGG-13/MobileV2). The results in both tables consistently demonstrate the superiority of \name over other knowledge distillation methods. Our method achieves state-of-the-art performance on CIFAR-100 on most teacher-student pairs, reaffirming its effectiveness and competitiveness in the field of knowledge distillation. 
\\

\noindent {\bf Results on ImageNet.}\quad
To assess the effectiveness of our method on large-scale datasets, we conduct experiment using ResNet-34 as the teacher and ResNet-18 as the student. Similar results on CIFAR-100 also occur in the ImageNet experiment. The results presented in Table~\ref{Tbl:ImageNet} demonstrate that our \name approach outperforms other existing distillation methods in terms of both Top-1 and Top-5 error rates. These findings highlight the efficacy of our method for learning on large-scale datasets.

It is worth noting that we compare our \name with the two existing graph-based methods (\ie, IRG and HKD). The key distinction lies in the construction of the graph structure. While the existing methods adopt an instance-wise graph construction, we construct the graph using patch tokens, thereby representing a more fine-grained and detailed structure.  By leveraging this refined relational knowledge, our \name consistently demonstrates superior performance when compared to the other two graph-based methods. These results further validate the effectiveness of our approach in acquiring and leveraging refined structural information, highlighting its potential in advancing graph-based knowledge distillation techniques.

\begin{table}[t]
\centering
\caption{Top-1 accuracies (\%) of student network  tiny/small DeiT~\cite{touvron2021training} on the ImageNet-1k validation set. ``-'' denotes that we deprecate the distillation token and conduct vanilla classification, and ``Hard-Label'' means that we distill the student by using the hard predictions of the teachers. 
}
\setlength{\tabcolsep}{3.0mm}{
\begin{tabular}{llcc}
    \toprule
    \multirow{2}{*}{Method} & \multirow{2}{*}{Student} & \multicolumn{2}{c}{Teacher}  \\
    \cmidrule{3-4}
    & & ResNet101 & CeiT-S\cite{yuan2021incorporating} \\
    \midrule
    - & DeiT-Tiny & 72.2 & 72.2 \\
    Hard-Label & DeiT-Tiny & 73.4 & 73.1 \\
    KD\cite{hinton2015distilling} & DeiT-Tiny & 74.8 & 74.5 \\
    HKD\cite{zhou2021distilling} & DeiT-Tiny & 75.2 & 74.8 \\
    \name & DeiT-Tiny & {\bf \color{blue}75.5} & {\bf \color{blue}75.4} \\
    \midrule
    - & DeiT-Small & 79.9 & 79.9 \\
    Hard-Label & DeiT-Small & 80.9 & 80.6 \\
    KD & DeiT-Small & 81.2 & 81.3 \\
    HKD & DeiT-Small & 81.5 & 81.3 \\
    \name & DeiT-Small & {\bf \color{blue} 82.0} & {\bf \color{blue}81.8} \\
    \bottomrule
\end{tabular}}
\label{Tbl:Vit results}
\end{table}

\subsubsection{Results with ViT-based Architectures}
We conduct additional experiments to evaluate the distillation performance on ViT-based networks. Following the training pipeline introduced in DeiT \cite{touvron2021training}, we employ DeiT-Tiny and DeiT-Small as the student models. To be consistent with the DeiT training procedure, we incorporate the distillation token and employ ResNet-101 and CeiT-Base \cite{yuan2021incorporating} as the teacher networks. The results are summarized in Table \ref{Tbl:Vit results}, from which we draw three key observations.

First, KD has proven to be beneficial for training ViTs, as all distillation results outperform the accuracy achieved without distillation. This finding demonstrates the effectiveness of knowledge transfer in improving the performance of ViT models. Furthermore, our observations indicate that employing  CNN-like networks as teachers for knowledge distillation yields superior results compared to using ViT models with the similar architectures. This performance disparity can be attributed to the local inductive bias introduced by CNN models, which enables them to effectively capture local patterns and features. Finally, our proposed \name consistently outperforms other distillation approaches in all the evaluated scenarios. These compelling results provide strong evidence of the effectiveness of our method in enhancing the performance of ViT models, highlighting its efficacy in knowledge distillation.

\begin{table}[t]
	\centering
 \caption{Ablation study of the proposed loss on CIFAR-100. Baseline denotes the primary CE classification loss on the student model without any distill. In other cases, the knowledge from pre-trained ResNet-32$\times$4 is used for distillation. The column `Token/Instance', indicates whether use the instance-level or token-level to construct relationship graph.}
	\resizebox{1\linewidth}{!}{
		\begin{tabular}{lccccccc}  
			\toprule
			\multirow{2}{*}{Module}& \multirow{2}{*}{Token/Instance}& \multicolumn{4}{c}{Losses}& \multirow{2}{*}{ResNet-8$\times$4} & \multirow{2}{*}{ShuffleV1}\\  
			&& $\mathcal{L}_\textrm{kd}$ & $\mathcal{L}_\textrm{inner}$ & $\mathcal{L}_\textrm{local}$ & $\mathcal{L}_\textrm{global}$ & & \\
			\midrule
			Baseline &- &-&-&-&-&72.51& 70.50 \\
			KD &Instance&\checkmark&-&-&-&74.12&74.00 \\
			w/o graph &Token&\checkmark&\checkmark&-&-&75.38&74.66\\
			w/o global&Token&\checkmark&\checkmark&\checkmark &-&75.91&75.68\\
            Instance-level &Instance&\checkmark&\checkmark&\checkmark&\checkmark&{76.12}&{76.01}\\
			\name &Token&\checkmark&\checkmark&\checkmark&\checkmark&{\bf \color{blue}76.42}&{\bf \color{blue}76.42}\\
			\bottomrule
	\end{tabular}}
	\label{tbl:loss comparison}
\end{table}

\begin{figure}
    \centering
    \includegraphics[scale=0.7]{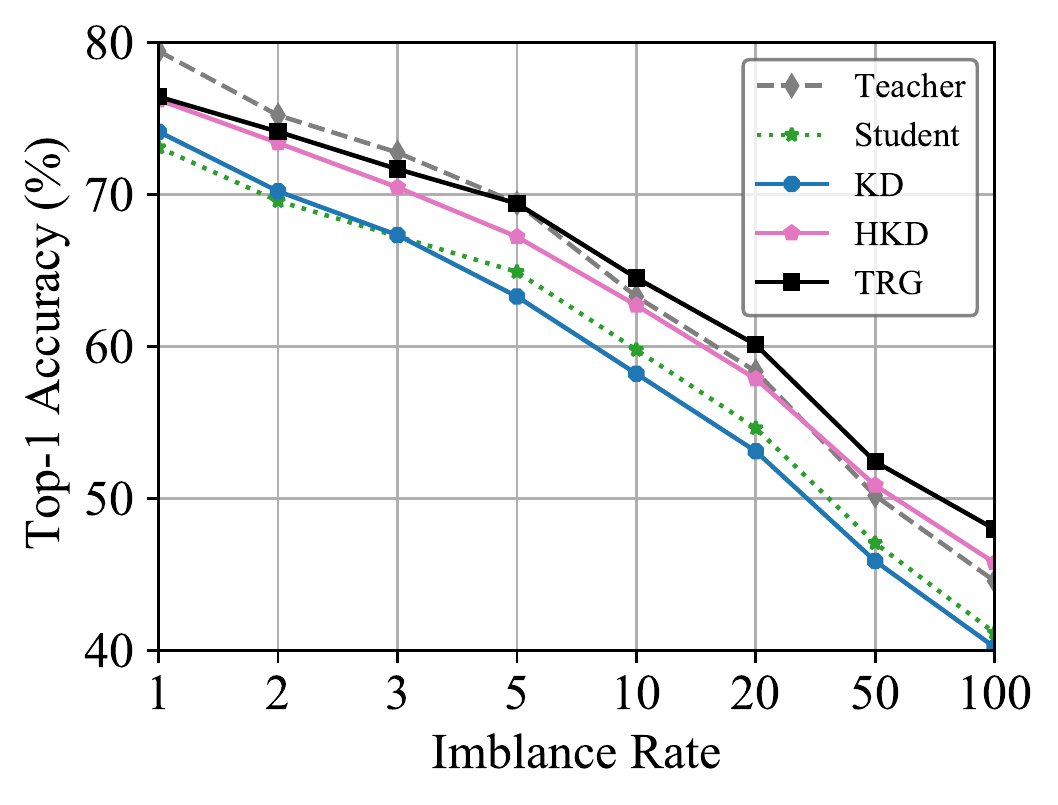}
    \caption{Top-1 accuracies (\%) on CIFAR-100-LT with different imbalance rates. We adopt ResNet-32$\times$4/ResNet-8$\times$4 as our teacher-student pair. The imbalance rates, defined as the frequency of the largest class divided by the smallest class.}
    \label{fig:imbalanced}
\end{figure}

\subsubsection{Results on Imbalanced-Data Scenarios}
{\noindent \bf Results on CIFAR-100-LT.}\quad
To demonstrate the effectiveness of our proposed approach, \name, in handling imbalanced data scenarios, we conduct a comparative evaluation of its classification performance on CIFAR-100-LT dataset with varying imbalanced rates. We compare our method against several baselines, including vanilla KD, HKD and plain teacher/student training. The results, as depicted in Fig.~\ref{fig:imbalanced}, reveal intriguing findings.
It is evident that when using plain training on either teacher network or student network, the performance is severely impacted by the long-tail effect. As the imbalance coefficient increases, the prediction accuracy decreases rapidly. Moreover,  KD loses its effectiveness compared to regular datasets, suggesting that a suboptimal teacher network can adversely affect the individual-level distillation process.

In contrast, both HKD and our proposed \name approach exhibit improved performance in predicting under imbalanced datasets, even surpassing the accuracy of the teacher's model. 
Notably, the curves representing our \name approach demonstrate the lowest sharpness and consistently outperform the teacher model in most scenarios. These results indicate that leveraging token-related information across instances can effectively mitigate the interference caused by the long-tail effect, which may severely affect the plain teacher training. Overall, these findings provide compelling evidence for the efficacy of our \name approach in enhancing the training process and mitigating the impact of imbalanced data distributions.

{\noindent \bf Results on ImageNet-LT.}\quad
In Table~\ref{Tbl:ImageNet}, we also present the performance comparison of our method with several existing approaches on ImageNet-LT. It can be observed that all experiments are impacted by the long-tail effect, resulting in a decrease in accuracy from 73.31\% to 50.11\% when using the teacher network. However, the relation-based methods (\ie, RKD, HKD and our \name) exhibit less vulnerability to the long-tail effect. Our method consistently maintains the highest performance and exhibits minimal decline on imbalanced datasets, surpassing even the performance of the teacher model. This outcome underscores the effectiveness of our approach in effectively addressing the challenges posed by imbalanced data scenarios.

\subsection{Ablation Studies}\label{ablation}
\subsubsection{t-SNE Visualization}
\begin{figure*} [t]
	\centering
	\subfloat[KD on ResNet8$\times$4]{
         \includegraphics[scale=0.26]{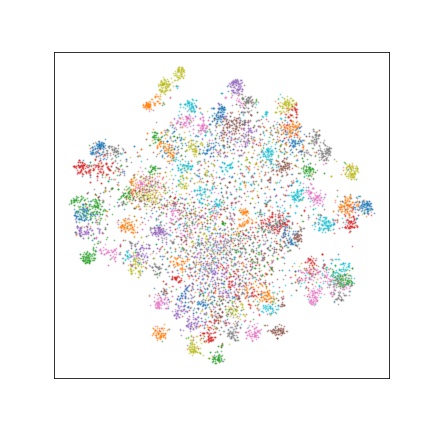}
    }
	\subfloat[IRG on ResNet8$\times$4]{
		\includegraphics[scale=0.26]{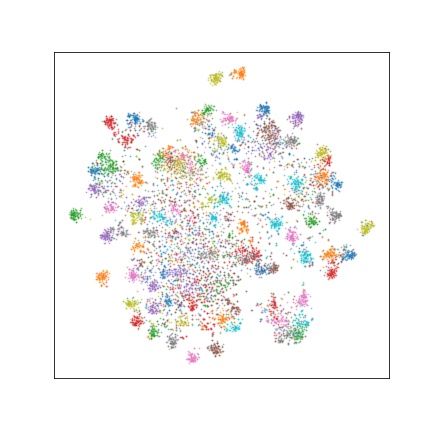}
		}
	\subfloat[HKD on ResNet8$\times$4]{
		\includegraphics[scale=0.26]{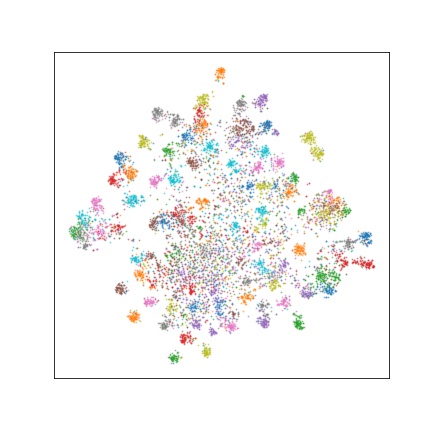}
		}
	\subfloat[Our \name on ResNet8$\times$4]{
		\includegraphics[scale=0.26]{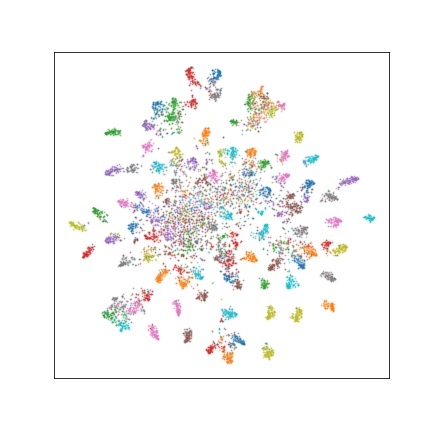}
		} \\
    
    \subfloat[KD on ShuffleV1]{\includegraphics[scale=0.26]{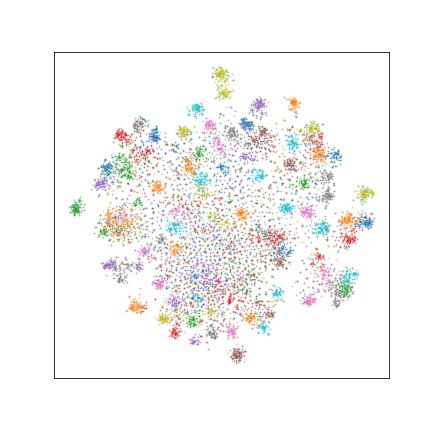}}
    \subfloat[IRG on ShuffleV1]{\includegraphics[scale=0.26]{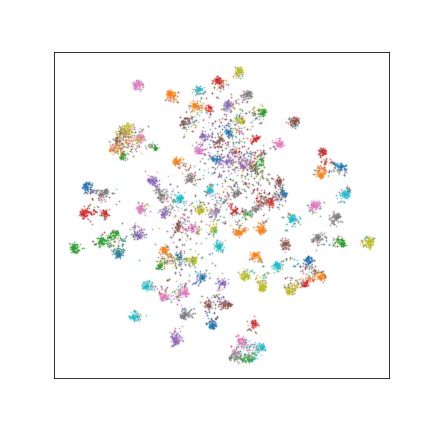}}
    \subfloat[HKD on ShuffleV1]{\includegraphics[scale=0.26]{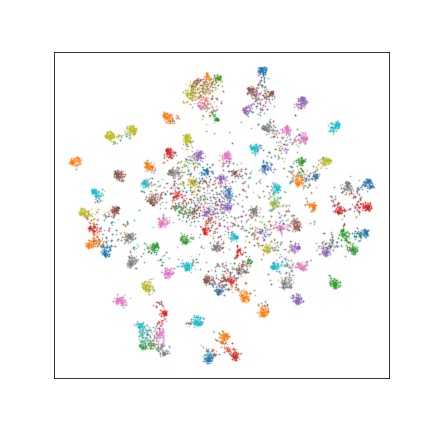}}
    \subfloat[Our \name on ShuffleV1]{\includegraphics[scale=0.26]{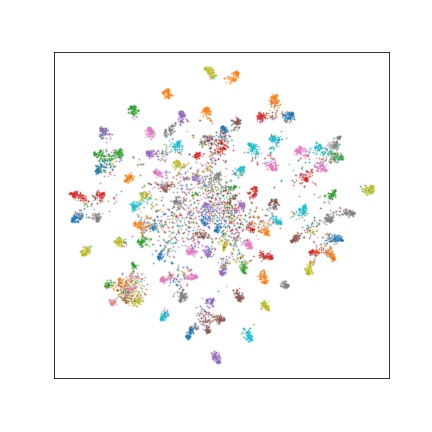}}

	\caption{t-SNE visualization of feature learned from ResNet-8$\times$4 
 (a-d) and ShuffleV1 (e-h) 
 on the CIFAR-100 dataset with the knowledge distilled from the pre-trained ResNet-32$\times$4. Here we use features before the classifier for visualization.}
	\label{fig:tsne} 
\end{figure*}

We present t-SNE visualizations of several existing distillation methods, including KD, two graph-based KD methods (\ie, IRG and HKD), and our proposed \name. Fig. \ref{fig:tsne} illustrates the visualizations obtained from these methods. Upon examination, it is evident that the representations generated by our \name exhibit superior separability when compared to the other distillation methods. Specifically, the t-SNE plot shows clear boundaries between different classes for our approach, whereas other methods demonstrate less distinct separations. This observation serves as evidence that our token-relationship approach enhances the robustness and discernibility of the feature representations. As a consequence, the classification performance is expected to benefit significantly from these improved representations.

\begin{figure}[htbp]
    \centering
	\includegraphics[width=0.65\linewidth]{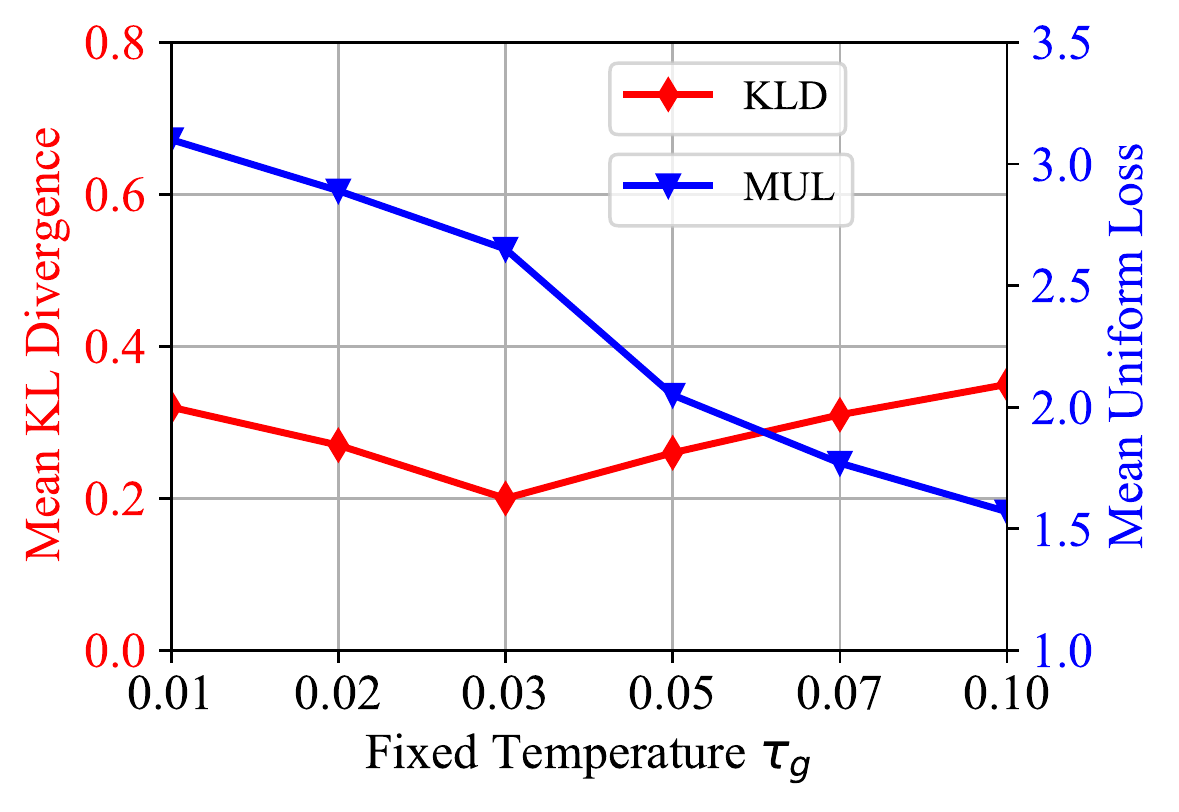}
		\caption{Mean KL Divergence and Mean Uniform Loss of feature selected from the distilled student ResNet-8$\times$4 (ResNet-32$\times$4 as the teacher) using several fixed temperature $\tau_g$. }
		\label{fig:uni}
\end{figure}
\begin{figure}[htbp]
    \centering
    \includegraphics[width=0.65\linewidth]{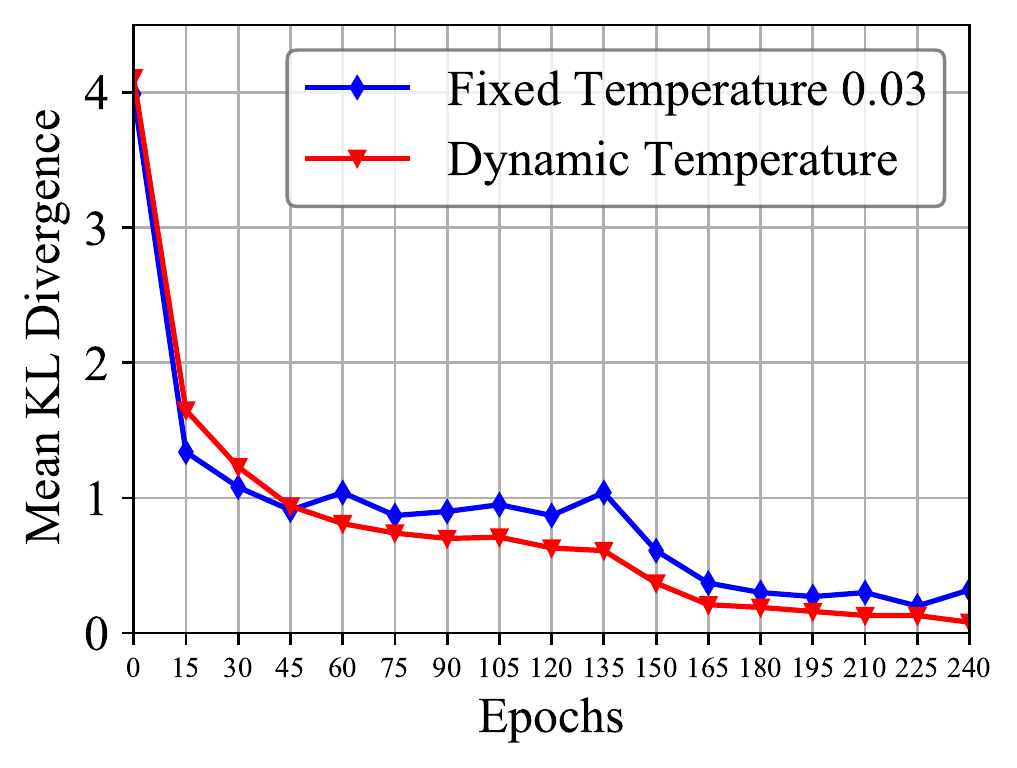}
		\caption{Mean KL Divergence using feature selected from the distilling student ResNet-8$\times$4 (use ResNet-32$\times$4 as the teacher) during the training process.}
		\label{fig:dyna_kl}
\end{figure}

\subsubsection{Ablation Study on Dynamic Temperature}
To assess the impact of the dynamic temperature strategy ($\tau_g$) employed in the graph contrastive loss (Eq.~\ref{eq:gcrd}), we compare our results with those obtained using fixed temperatures. Our evaluation is conducted on the CIFAR-100 dataset, where we utilize the student ResNet-8$\times$4, distilled from the pre-trained ResNet-32$\times$4. Additionally, we also compute two additional metrics. The first metric is the Mean KL Divergence (KLD), which quantifies the discrepancy between the penultimate features of the student and its corresponding teacher. The second metric, called Mean Uniformity Loss (MUL), quantifies the mean Euclidean distance between tokens in relation to other tokens as follows:
\begin{equation}
    \mathrm{MUL} = \frac{1}{|{\mathbf{T}}|}\sum_{i,j\in\mathbf{T}}(\mathbf{T}_i - \mathbf{T}_j)^2,
\end{equation}
where $|{\mathbf{T}}|$ is the number of tokens. 
MUL serves as a measure of the dispersity of the learned feature space. 

The evaluation results are presented in Fig.~\ref{fig:uni} and Fig.~\ref{fig:dyna_kl}. From Fig.~\ref{fig:uni}, it is evident that the MUL increases as the temperature $\tau_g$ reduces. This finding aligns with the proposition of Wang \etal~\cite{wang2020understanding}, suggesting that a lower temperature in the contrastive loss tends to focus more on hard samples, resulting in a more dispersed feature distribution. However, it is important to note that a dispersed distribution does not necessarily imply better classification performance. In Fig.~\ref{fig:uni}, we observe that the best result, as manifested by the KLD curve (which exhibits a similar pattern to the accuracy curve), occurs at a relatively moderate temperature of 0.03. This suggests that an optimal balance between concentration and dispersion is crucial for achieving optimal performance.

We compare the performance of our dynamic temperature strategy with the best fixed temperature setting ($\tau_g=0.03$, lowest KLD shown in Fig.~\ref{fig:uni}). Fig.~\ref{fig:dyna_kl} illustrates the KLD 
during the training process. We observe distinct patterns in the behavior of the KLD under fixed and dynamic temperature settings. When using a fixed temperature, the KLD initially decreases rapidly but reaches at a plateau and exhibits fluctuations before $150$ epochs. At epoch 150, the learning rate is reduced, leading to a drop in the Mean KL Divergence; however, eventually it settles at a relatively high value. In contrast, our dynamic temperature approach employs a higher initial temperature, enabling the contrastive loss to focus on coarse-grained classification at the early training stage. As the epochs progress, the temperature dynamically decreases, shifting the focus towards contrastive learning of hard samples. This dynamic temperature strategy prevents the observed oscillation in KLD under the fixed temperature, facilitating a continuous decrease. Consequently, our approach achieves a relatively lower KLD value, indicating improved classification performance.
Notably, our dynamic temperature strategy outperforms all fixed temperature strategies, highlighting the efficacy of our approach's dynamic temperature adjustment. The results demonstrate the importance of adaptively adjusting the temperature parameter for effective knowledge distillation.

 \subsubsection{Ablation Study on Different Losses}
 We conduct another ablation study on the proposed loss using the CIFAR-100 dataset, and the results are presented in Table~\ref{tbl:loss comparison}. In the table, `KD' indicates that we solely utilize vanilla KD loss $\mathcal{L}_\textrm{logit}$, and `w/o graph' denotes the reduction of graph-based loss for knowledge distillation. We observe that in both networks, all the proposed losses manifest its part in transfer knowledge effectively. Furthermore, we compare our results with a control group that employs all the proposed losses but at the instance-level. As shown in Table~\ref{tbl:loss comparison}, when we switch from token-wise graph to instance-level, the accuracy 
 drops by 0.30\% and 0.41\% for the two networks, respectively. A conclusion can be drawn that token-wise knowledge plays a crucial role in the distillation process, demonstrating its benefits over instance-level knowledge.

 \subsubsection{Ablation Study on Batch Size}
 \begin{table}[t]
\centering
\caption{Effect of batch size on distillation (100-epoch ImageNet evaluation accuracy with knowledge distill from ResNet34 to ResNet18).
}
\begin{tabular}{c|cccc}
\toprule
Batch size  & 128 & 256 & 512 & 1024  \\
\midrule
Acc. (\%)  & 70.93 & 71.09 & {\bf \color{blue} 71.31} & 71.01  \\ 
\bottomrule
\end{tabular}
\label{tbl:batch}
\end{table}

Table \ref{tbl:batch} presents the impact of different batch sizes on the distillation performance. It can be observed that increasing the batch size from 128 to 512 leads to improved accuracy. This improvement can be attributed to the fact that larger batch sizes allow for the inclusion of more instances, which in turn enables the construction of a more representative graph with greater number of tokens.

Nevertheless, 
it is important to note that excessively large batch sizes do not lead to favorable distillation results. In particular, when using a batch size of 1024, the performance is lower compared to the case with a batch size of 512. This phenomenon can be attributed to the challenges associated with constructing a large-scale graph. As the batch size increases, it becomes increasingly difficult to effectively capture the relationships and interactions among a larger number of tokens. This observation further emphasizes the significance of utilizing a random sampling strategy to maintain the essential properties of the graph construction process.

\label{05Exp}

\section{Conclusion}
In this work, we delved into an important yet relatively unexplored aspect of knowledge distillation (KD). We argued that the token-wise relational information not only encompasses the semantic context within individual images but also captures latent token-level relationship information across instances. Motivated by this insight, we proposed a novel relation-based distillation method called \name, which constructs a graph at the token level to facilitate the transmission of relational knowledge among tokens. Through extensive experiments on various visual classification tasks, including those involving imbalanced datasets, our \name method has achieved state-of-the-art performance. The results strongly support the effectiveness and efficiency of \name in distilling knowledge. Notably, in the context of imbalanced-data recognition tasks, our method has surpassed the teacher's model in terms of classification performance, indicating its ability to alleviate the long-tail effect. This observation highlights the potential of token-level relational knowledge in addressing the challenges posed by imbalanced data distributions. Building on the success of \name in classification tasks, it is expected that our method can also be applied to more complex visual tasks such as object detection and semantic segmentation, which we may study in the future. 
\label{06Con}

\section*{Acknowledgments}
This work is supported by the National Natural Science Foundation of China (No. U22B2017).





\ifCLASSOPTIONcaptionsoff
  \newpage
\fi



\bibliographystyle{IEEEtran}
\bibliography{TBD}
%



%

\begin{IEEEbiography}[{\includegraphics[width=1in,height=1.25in,clip,keepaspectratio]{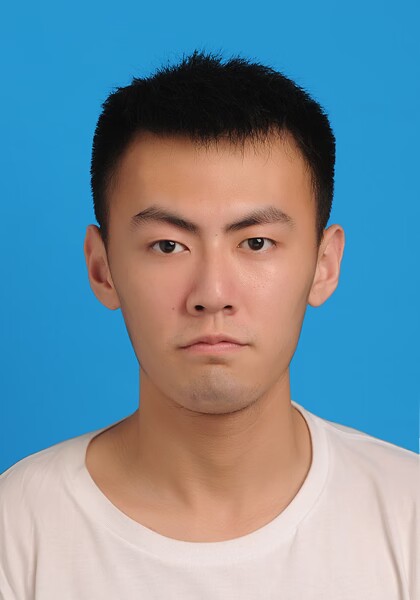}}]{Shuoxi~Zhang}
received the master’s degree in pure mathematics from Wuhan University, China, in 2018. He is currently pursuing the Ph.D. Degree with the School of Computer Science, Huazhong University of Science and Technology. His research interests include unsupervised learning, knowledge transfer and network architecture design.
\end{IEEEbiography}
\begin{IEEEbiography}[{\includegraphics[width=1in,height=1.25in,clip,keepaspectratio]{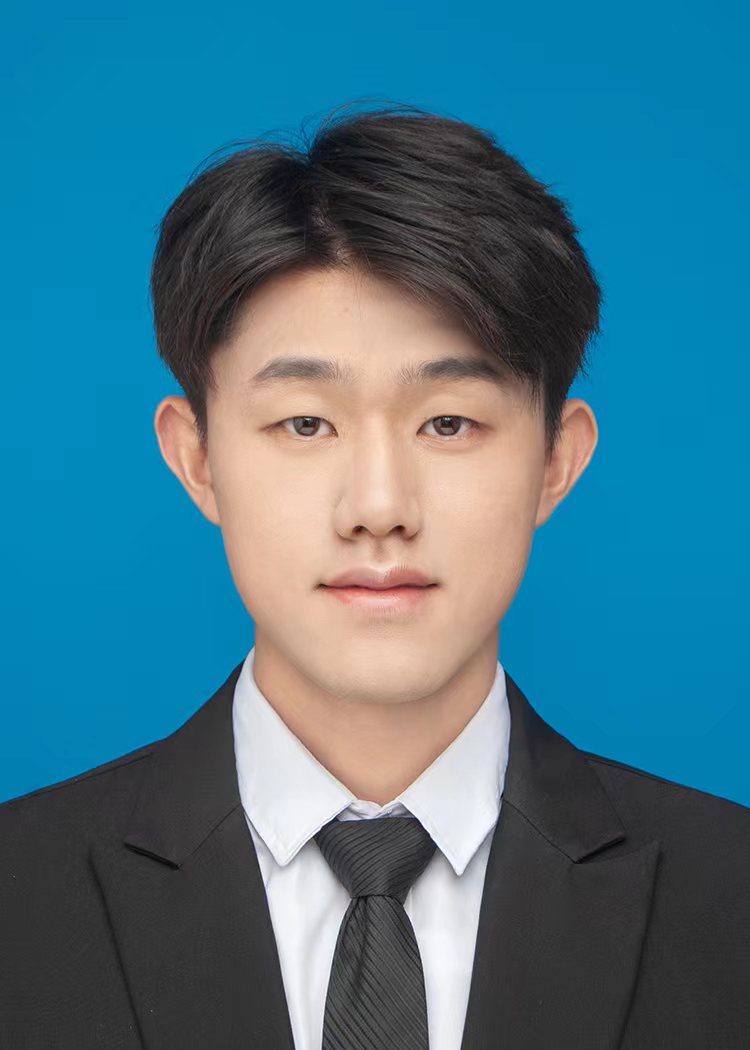}}]{Hanpeng~Liu}
received the bachelor’s degree in computer science from Huazhong University of Science and Technology, China, in 2022. He is currently pursuing the Ph.D. Degree with the School of Computer Science, Huazhong University of Science and Technology. His research interests include knowledge distillation, self-supervised learning and adversarial learning.
\end{IEEEbiography}
\begin{IEEEbiography}[{\includegraphics[width=1in,height=1.25in,clip,keepaspectratio]{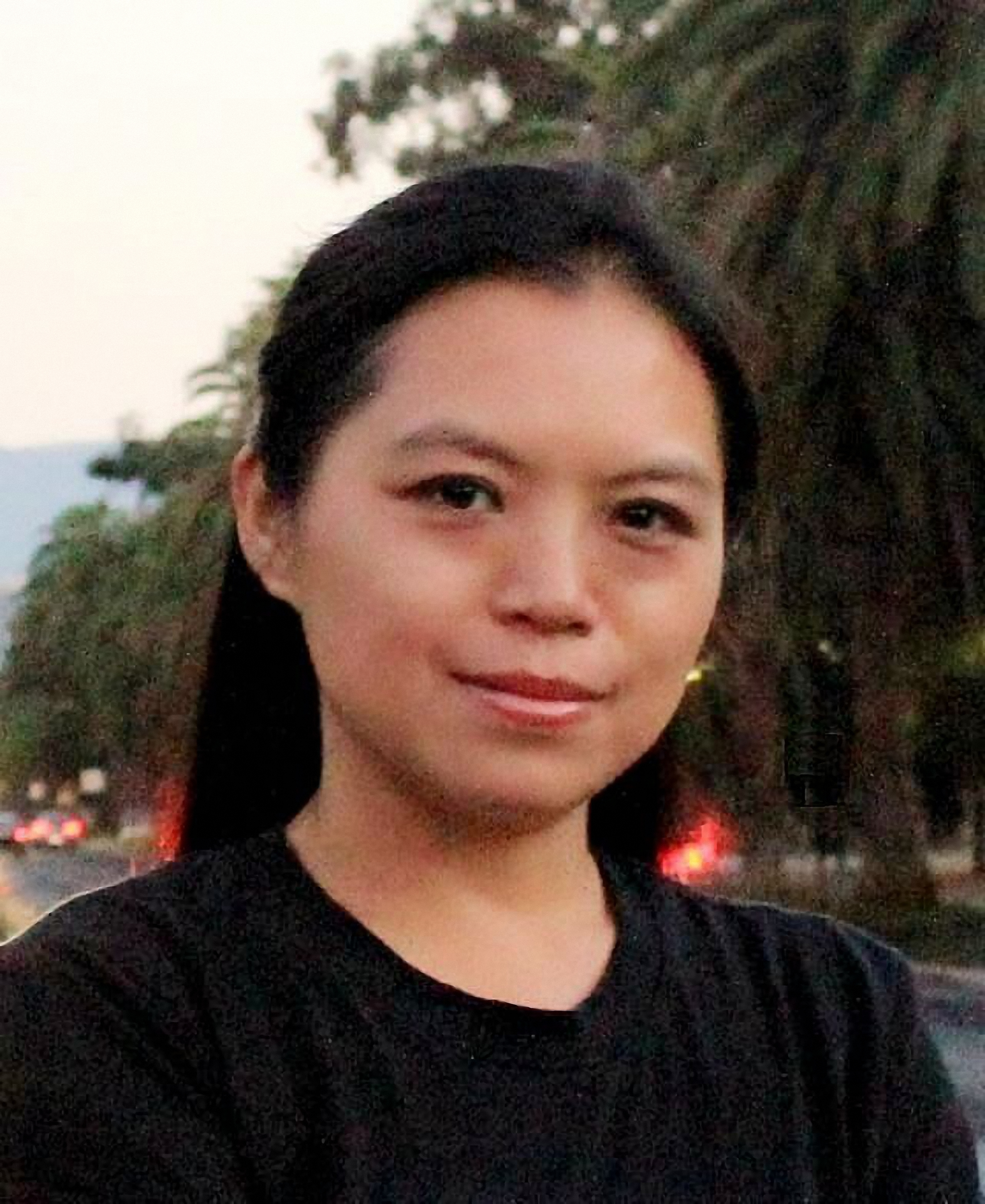}}]{Kun He} (SM18) received the Ph.D. degree in system engineering from Huazhong University of Science and Technology, Wuhan, China, in 2006. She is currently a Professor in School of Computer Science and Technology, Huazhong University of Science and Technology, Wuhan, China. She had been with the Department of Management Science and Engineering at Stanford University in 2011-2012 as a visiting researcher. 
She had been with the department of Computer Science at Cornell University in 2013-2015 as a visiting associate professor, in 2016 as a visiting professor, and in 2018 as a visiting professor.  She was honored as a Mary Shepard B. Upson visiting professor for the 2016-2017 Academic year in Engineering, Cornell University, New York. Her research interests include adversarial learning, representation learning, social network analysis, and combinatorial optimization.  
\end{IEEEbiography}


\vfill


\end{document}